\documentclass{article} 
\usepackage{iclr2024_conference,times}

\usepackage{amsmath,amsfonts,bm}
\usepackage{graphicx}
\usepackage{subcaption}

\def\eqref#1{equation~\ref{#1}}

\def\1{\bm{1}}

\DeclareMathAlphabet{\mathsfit}{\encodingdefault}{\sfdefault}{m}{sl}
\SetMathAlphabet{\mathsfit}{bold}{\encodingdefault}{\sfdefault}{bx}{n}

\usepackage{hyperref}
\usepackage{url}
\usepackage{bm}
\usepackage{amssymb}
\usepackage{amsmath}
\usepackage{algorithm}
\usepackage{algpseudocode}
\usepackage{booktabs}

\definecolor{reviewA}{HTML}{01A2FF}

\newcommand{\bbb}{\noindent\textbf}
\newcommand{\aname}{INT-FlashAttention}

\title{INT-FlashAttention: Enabling Flash Attention for INT8 Quantization}

\author{Shimao Chen\textsuperscript{1}, Zirui Liu\textsuperscript{1}, Zhiying Wu\textsuperscript{2}, Ce Zheng\textsuperscript{3}, Peizhuang Cong\textsuperscript{1},\\ \textbf{Zihan Jiang\textsuperscript{1}, Yuhan Wu\textsuperscript{1}, Lei Su\textsuperscript{2}, Tong Yang\textsuperscript{1}\thanks{Corresponding author: Tong Yang (yangtong@pku.edu.cn)}}\\
\textsuperscript{1} Peking University, 
\textsuperscript{2} Baichuan Inc,
\textsuperscript{3} Beihang University.
}

\iclrfinalcopy 
\begin{document}

\maketitle

\begin{abstract}

As the foundation of large language models (LLMs), self-attention module faces the challenge of quadratic time and memory complexity with respect to sequence length. 
FlashAttention accelerates attention computation and reduces its memory usage by leveraging the GPU memory hierarchy.
A promising research direction is to integrate FlashAttention with quantization methods.
This paper introduces \aname{}, the first INT8 quantization architecture compatible with the forward workflow of FlashAttention, which significantly improves the inference speed of FlashAttention on Ampere GPUs.
We implement our \aname{} prototype with fully INT8 activations and general matrix-multiplication (GEMM) kernels, making it the first attention operator with fully INT8 input.
As a general token-level post-training quantization framework, \aname{} is also compatible with other data formats like INT4, \textit{etc.}
Experimental results show \aname{} achieves 72\% faster inference speed and 82\% smaller quantization error compared to standard FlashAttention with FP16 and FP8 data format. 
All related codes are open-sourced \footnote{\url{https://github.com/INT-FlashAttention2024/INT-FlashAttention}}. 
\end{abstract}
\section{Introduction}
Large language models (LLMs), such as GPT \cite{achiam2023gpt} and Llama \cite{touvron2023llama}, have achieved significant breakthroughs across various domains. 
Self-attention module is the foundation of LLMs, allowing them to capture dependencies between different tokens in a sequence \cite{vaswani2017attention}.
However, the computation of self-attention module entails quadratic time and memory complexity with respect to the sequence length, which hinders the scaling of LLMs to longer contexts.
To address this bottleneck, Dao et al. propose excellent FlashAttention \cite{flashattention1}, which exploits the GPU memory hierarchy to design a tiling strategy to speed up the attention computation and reduce memory from quadratic to linear w.r.t. sequence length. 


Another popular method of improving the computational performance and memory usage of LLMs is to reduce the bit size of the floating-point data used in these models, which is a technique known as model quantization. 
Modern hardware incorporates advanced Tensor Cores to support efficient general matrix-multiplication (GEMM) for FP16, FP8 \cite{micikevicius2022fp8}, and INT8 \cite{van2023fp8}.  
Quantization methods make full use of these dedicated computing units to improve the spatiotemporal efficiency of LLMs by compressing the parameters and activations into FP8, INT8, and even ternary format. 
Existing quantization methods can be broadly divided into two categories: quantization during training and post-training quantization. 
This paper focuses on the latter, utilizing quantization to enhance LLM inference efficiency.




Integrating FlashAttention with quantization methods is a promising research direction. 
Designed for the most advanced NVIDIA Hopper GPUs, the latest FlashAttention-3 \cite{flashattention3} already supports FP8 data format. 
Unfortunately, NVIDIA Ampere series GPUs, such as A100, do not provide hardware support for FP8. 
While Hopper architecture significantly enhances both computational power and energy efficiency, existing Ampere architecture still holds a substantial share in the data center GPU market. 
According to available data, A100 GPUs are still estimated to contribute approximately 20\% of the total compute power in accelerated supercomputers \cite{ampere}. This dominance is largely due to their extensive deployment across numerous data centers worldwide.
Given that the Ampere architecture provides computational support for INT8 data format, this paper aims to develop a fully INT8 version of FlashAttention, significantly improving the inference speed of FlashAttention on Ampere GPUs compared to the basic FlashAttention with FP16. 

Towards the above design goal, we propose \aname{}, a novel token-level
post-training quantization architecture designed to align with the computation workflow of FlashAttention. 
We implement our \aname{} prototype with INT8-type $\mathbf{Q}$, $\mathbf{K}$, and $\mathbf{V}$ matrices. 
We use INT8 general matrix-multiplication (GEMM) kernels to replace all matrix multiplications during inference, thus significantly improving inference speed and saving energy. 
To the best of our knowledge, we are the first to develop the attention operator with fully INT8 inputs.
By preserving token-level information, \aname{} offers a substantial accuracy improvement over existing tensor-level FP8 methods \cite{flashattention3}.
Last but not least, our token-level quantization method is not limited to INT8 format, which can also be adapted to other data formats like INT4, and \textit{etc}.
Experimental results show that the INT8 version of \aname{} achieves 72\% faster inference speed compared to FlashAttention-FP16  and up to 82\% smaller quantization error compared to FlashAttention-FP8. 

This paper makes the following contributions:
\begin{itemize}
    \item We propose \aname{}, a token-level post-training quantization architecture that can be seamlessly integrated with the forward workflow of FlashAttention. 
    \item We implement the INT8 version of our \aname{} prototype, which is the first attention operator with fully INT8 input (to the best of our knowledge). 
    \item We conduct experiments evaluating the performance of \aname{}, showing that it achieves significantly faster inference speed and higher quantization accuracy than baseline solutions. 
\end{itemize}
\section{Background and Related Work}

\subsection{Standard Attention}

Given input sequence $\mathbf{Q,K,V}\in \mathbb{R}^{N\times d}$, where $N$ is the sequence length and $d$ is the head dimension, standard attention computes the attention output $\mathbf{O} \in \mathbb{R}^{N\times d}$:

$$\mathbf{S} = \mathbf{Q}\mathbf{K}^T \in \mathbb{R}^{N\times N}, \; \mathbf{P} = softmax(\mathbf{S}) \in \mathbb{R}^{N\times N}, \; \mathbf{O} = \mathbf{P}\mathbf{V} \in \mathbb{R}^{N\times d},$$

where softmax is applied row-wise. Here, we call $\mathbf{S}$ \textit{attention score matrix} and call $\mathbf{P}$ \textit{attention weight matrix}. 

Standard attention implements the intermediate matrices $\mathbf{S}$ and $\mathbf{P}$ to HBM, which takes $O(N^2)$ memory. 
During forward computation, standard attention implementation first loads the entire inputs from GPU high bandwidth memory (HBM), calculates $\mathbf{S} = \mathbf{Q}\mathbf{K}^T$ and writes $\mathbf{S}$ back to HBM. 
Then it loads $\mathbf{S}$ from HBM, calculates $\mathbf{P} = softmax(\mathbf{S})$, and writes $\mathbf{P}$ back to HBM. 
It finally calculates $\mathbf{O} = \mathbf{P}\mathbf{V}$ to get the final results. 
We can see that in the circumstance above, most of the operations are bounded by the HBM bandwidth, and thus the large number of memory accesses dominates the time taken by attention operations.

\subsection{FlashAttention Families}

To speed up attention on hardware accelerators with hierarchical memory, FlashAttention \cite{flashattention2} proposes to use the tiling techniques to reduce memory reads/writes and fuse the attention operations into a single kernel. 
Specifically, FlashAttention divides the input sequences into smaller blocks. During the forward computation, it first loads blocks of inputs from HBM to SRAM, computes attention for each block and then updates the output without writing the large intermediate matrices $\mathbf{S}$ and $\mathbf{P}$ back to HBM. 
Since the softmax function in the attention mechanism needs to be applied to the entire row, FlashAttention employs an online softmax method \cite{milakov2018online,rabe2021self} to split the computation into blocks and finally rescale the output of each block to obtain the correct result.
Built on FlashAttention, FlashAttention-2 \cite{flashattention2} modifies the design of the inner loop in the forward pass and introduces two tweaks to reduce non-matmul FLOPs, thereby improving GPU occupancy. 
FlashAttention-3 \cite{flashattention3} further leverages Tensor Cores and TMA asynchrony to optimize the FlashAttention implementation on Hopper GPUs. 
In particular, it utilizes block quantization and the hardware support for FP8 low-precision to implement an FP8 version of FlashAttention on Hopper GPUs.

\subsection{Model Quantization}

\bbb{LLM quantization.}
Recent years have witnessed a great surge in applying various quantization methods to reduce the memory and energy consumption of LLMs. 
Quantization methods compress the parameters from standard FP32 to continuous FP16 \cite{flashattention1}, FP8 \cite{flashattention3,lee2024fp8}, discrete INT8 \cite{dettmers2022gpt3}, and even ternary formats \cite{chen2024ternaryllm}. 
Many studies suggest that INT8 consumes significantly less memory for loading model weights and requires less energy than the FP8 and FP16 counterparts \cite{dally2015high,van2023fp8}. 
The naive LLM quantization approaches adopt tensor-level quantization \cite{zhou2024survey}. 
However, many studies found that the weight distributions vary significantly across different tokens, and the existence of activation outliers makes LLM difficult to quantize at whole-tensor level \cite{tao2022compression,xu2024empowering}.  
As a result, many works adopted token-level or block-level quantization methods to improve model accuracy \cite{zhou2024survey,li2024evaluating}.
The most recent FlashAttention-3 \cite{flashattention3} adopts a block-level FP8 quantization method.
However, to the best of our knowledge, there is no existing work that integrates token-level quantization with FlashAttention, neither is there a version of FlashAttention that supports INT8 data format.

\bbb{Post-training quantization.}
There is a line of post-training quantization (PTQ) strategies for effectively reducing model sizes and improving the inference speed of LLMs. 
Some work design custom quantization blocks to improve quantization accuracy \cite{dettmers2022gpt3,li2021brecq}. 
Other works use feature segmentation strategies to protect the outlier features so as to reduce the quantization error \cite{shang2023pb,dettmers2023spqr}. 
GPTQ \cite{frantar2022gptq} designs a more accurate one-shot quantization framework based on approximate second-order information \cite{frantar2022optimal}, achieving good accuracy in extreme low-bit quantization regime (2-bit).  
Subsequent works propose to identify and select salient weights and preserve their information through scaling transformations \cite{lee2023owq} or residual approximation \cite{huang2024billm}.




\section{The \aname{} Architecture}

Developed based on FlashAttention \cite{flashattention1}, \aname{} implements the $\mathbf{Q}$, $\mathbf{K}$, and $\mathbf{V}$ matrices in fully INT8 format (colored green in Figure~\ref{fig:alg_int8_attention}) with token-level quantization. 
We use the INT8 general matrix multiplication (GEMM) kernel to replace all matrix multiplications in FlashAttention (originally in FP16/FP8 format).
Due to the efficient INT8 compression, \aname{} can read larger blocks from HBM per iteration compared to basic FlashAttention with FP16 format. 
Additionally, the INT8 matrix multiplication operators in \aname{} also outperform their floating-point counterparts.
As a result, \aname{} achieves notable improvement in inference speed compared to basic FlashAttention. 
Furthermore, with per-token quantization, \aname{} offers better inference accuracy than the FP8 version of FlashAttention-3 \cite{flashattention3} with tensor-level quantization.

In this section, we first introduce the workflow of the online softmax operation in Section~\ref{subsec:softmax}, which forms the foundation of both FlashAttention and \aname{}. 
Then we explain how \aname{} seamlessly integrates INT8 quantization into the online softmax workflow in Section~\ref{subsec:superflash}.

\subsection{Preliminary of Online Softmax}
\label{subsec:softmax}

\begin{figure*}[tb]
    \centering
    \includegraphics[trim=0.1cm 1cm 0cm 0cm, clip,width=1\linewidth]{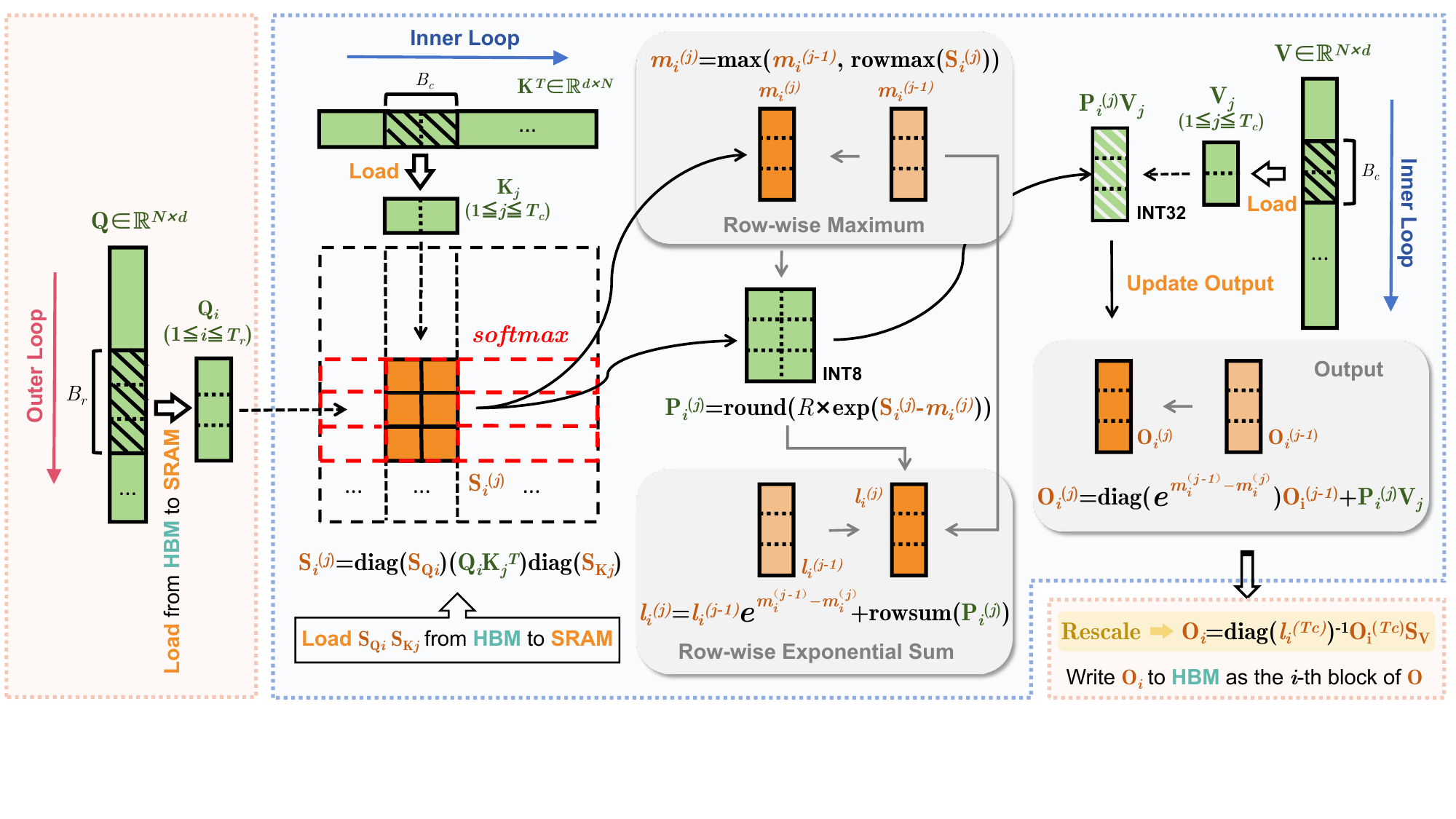}
    \caption{A single iteration in the inner loop of \aname{} forward pass (Algorithm~\ref{code:algo}), where green color represents INT8 data format and orange color represents FP32 data format.}
    \label{fig:alg_int8_attention}
\end{figure*}

We first describe the workflow of the online softmax method \cite{milakov2018online,rabe2021self} in detail, which is the base of FlashAttention \cite{flashattention1} and our \aname{}.  
As shown in Figure~\ref{fig:alg_int8_attention} and Algorithm~\ref{code:algo}, in the forward workflow of FlashAttention (also our \aname{}), we iterate over blocks of the input $\mathbf{Q}$ and $\mathbf{K}/\mathbf{V}$ matrices. 
For each iteration in the outer loop (line 4-5 in Algorithm~\ref{code:algo}), we load a block $\mathbf{Q}_i$ from HBM, and then perform the inner loop over blocks of the $\mathbf{K}$ and $\mathbf{V}$ matrices.
In each iteration of the inner loop (line 7-8 in Algorithm~\ref{code:algo}), we load the blocks $\mathbf{K}_j$ and $\mathbf{V}_j$ from HBM and compute $\mathbf{S}_i^{(j)} = \mathbf{Q}_i \mathbf{K}_j^T$. 
In standard attention, a row-wise softmax operation is required on the entire attention matrix $\mathbf{S}_i = \left[\mathbf{S}_i^{(1)}\; \mathbf{S}_i^{(2)} \; \cdots \; \mathbf{S}_i^{(T_c)} \right]$.
The online softmax method achieves the same result as standard attention by scaling the output of each inner loop iteration with the appropriate normalization factor.

For simplicity, consider there are only one-row block and two column blocks in the attention matrix $\mathbf{S} = \left[\mathbf{S}^{(1)} \; \mathbf{S}^{(2)} \right]$ where $\mathbf{S}^{(1)}$, $\mathbf{S}^{(2)} \in \mathbb{R}^{B_r \times B_c}$.  
We want to compute row-wise softmax on $\mathbf{S}$ and multiply the result $\mathbf{P} = softmax(\mathbf{S})$ with the value 
$\mathbf{V} = \left[ \begin{array}{c}
\mathbf{V}^{(1)} \\
\mathbf{V}^{(2)}
\end{array} \right]$ where $\mathbf{V}^{(1)}$, $\mathbf{V}^{(2)} \in \mathbb{R}^{B_c \times d}$.
Standard softmax computes as follows.
\begin{align*}
m &= \max(\text{rowmax}(\mathbf{S}^{(1)}), \text{rowmax}(\mathbf{S}^{(2)})) \in \mathbb{R}^{B_r} \\
l &= \text{rowsum}(e^{\mathbf{S}^{(1)} - m}) + \text{rowsum}(e^{\mathbf{S}^{(2)} - m}) \in \mathbb{R}^{B_r} \\
\mathbf{P} &= 
\begin{bmatrix} 
\mathbf{P}^{(1)} & \mathbf{P}^{(2)} 
\end{bmatrix}
= \text{diag}(l)^{-1} 
\begin{bmatrix} 
e^{\mathbf{S}^{(1)} - m} & e^{\mathbf{S}^{(2)} - m}
\end{bmatrix} 
\in \mathbb{R}^{B_r \times 2B_c} \\
\mathbf{O} &= 
\begin{bmatrix} 
\mathbf{P}^{(1)} & \mathbf{P}^{(2)}
\end{bmatrix}
\begin{bmatrix}
\mathbf{V}^{(1)} \\
\mathbf{V}^{(2)}
\end{bmatrix}
= \text{diag}(l)^{-1} e^{\mathbf{S}^{(1)} - m} \mathbf{V}^{(1)} + e^{\mathbf{S}^{(2)} - m} \mathbf{V}^{(2)} 
\in \mathbb{R}^{B_r \times d}
\end{align*}

Online softmax computes local softmax within each block $\mathbf{S}^{(i)}$ and finally rescales the results to get the same output as standard softmax. 
The following formulas describe the online softmax implemented in FlashAttention-2 \cite{flashattention2}, which also forms the foundation of our \aname{}. 
We annotate these formulas with their corresponding line numbers in the pseudocode of Algorithm~\ref{code:algo}.
\begin{align*}
m^{(1)} &= \text{rowmax}(\mathbf{S}^{(1)}) \in \mathbb{R}^{B_r} \\
l^{(1)} &= \text{rowsum}(e^{\mathbf{S}^{(1)} - m^{(1)}}) \in \mathbb{R}^{B_r} \\
\widetilde{\mathbf{O}}^{(1)} &= e^{\mathbf{S}^{(1)} - m^{(1)}} \mathbf{V}^{(1)} \in \mathbb{R}^{B_r \times d} \\
m^{(2)} &= \max(m^{(1)}, \text{rowmax}(\mathbf{S}^{(2)})) = m &\text{(Line 10)}\\
l^{(2)} &= e^{m^{(1)} - m^{(2)}} l^{(1)} + \text{rowsum}(e^{\mathbf{S}^{(2)} - m^{(2)}})  = l &\text{(Line 12)} \\
\widetilde{\mathbf{O}}^{(2)} &= \text{diag}(e^{m^{(1)} - m^{(2)}}) \widetilde{\mathbf{O}}^{(1)} + e^{\mathbf{S}^{(2)} - m^{(2)}} \mathbf{V}^{(2)} = e^{\mathbf{S}^{(1)} - m} \mathbf{V}^{(1)} + e^{\mathbf{S}^{(2)} - m} \mathbf{V}^{(2)} &\text{(Line 13)}\\
\mathbf{O}^{(2)} &= \text{diag}(l^{(2)})^{-1} \widetilde{\mathbf{O}}^{(2)} = \mathbf{O} &\text{(Line 16)}.
\end{align*}

From the formulas above, we can see that during the inner loop iterations, by real-time maintaining the current row-wise maximum values $m^{(j)}$ and the sum of exponentials $l^{(j)}$, and by finally rescaling the output, online softmax can attain the same output as standard softmax. 

\begin{algorithm}[ht]
\caption{\aname{} Attention Forward}\label{alg:cap}
\label{code:algo}

\begin{algorithmic}[1]
\Require 
Matrices $\mathbf{Q,K,V}\in \mathbb{I}_8^{N\times d}$ ($\mathbb{I}_8 := [-128,127] \cap \mathbb{Z}$ is the range of INT8), 
scalers $\mathbf{S_Q,S_K}\in \mathbb{R}^{N}$, 
$\mathbf{S_V}\in \mathbb{R}$ in HBM, 
block sizes $B_c, B_r$, INT8 maximum $R=127$.
\State
Divide $\mathbf{Q}$ into $T_r=\left\lceil \frac{N}{B_r}\right\rceil$ blocks $\mathbf{Q}_1,\dots,\mathbf{Q}_{T_r}$, each have size $B_r\times d$, and divide $\mathbf{K,V}$ into $T_c=\left\lceil \frac{N}{B_c} \right\rceil$ blocks $\mathbf{K}_1,\dots,\mathbf{K}_{T_c}$ and $\mathbf{V}_1,\dots,\mathbf{V}_{T_c}$, each have size $B_c\times d$.
\State
Divide output $\mathbf{O}\in \mathbb{R}^{N\times d}$ into $T_r$ blocks $\mathbf{O}_1,\dots, \mathbf{O}_{T_r}$, each have size $B_r\times d$.
\State
Divide $\mathbf{S_Q}$ into $T_r$ blocks ${\mathbf{S_Q}}_1,\dots{\mathbf{S_Q}}_{T_r}$, each have size $B_r$, 
and divide $\mathbf{S_K}$ into $T_c$ blocks ${\mathbf{S_K}}_1,\dots{\mathbf{S_K}}_{T_c}$, each have size $B_c$, 
\For{$1\leq i \leq T_r$}
    \State Load $\mathbf{Q}_i$ and $\mathbf{{S_Q}}_i \in \mathbb{R}^{B_r}$ from HBM to on-chip SRAM.
    \State On chip, initialize $\mathbf{O}_i^{(0)}=(0)_{B_r\times d}, 
            l_i^{(0)}=(0)_{B_r}, m_i^{(0)}=(-\infty)_{B_r}$
    \For{$1\leq j \leq T_c$}
        \State Load $\mathbf{K}_j, \mathbf{V}_j, \mathbf{{S_K}}_j \in \mathbb{R}^{B_c}$ from HBM to on-chip SRAM.
        \State On chip, compute $\mathbf{S}_i^{(j)}= \text{diag}\left( \mathbf{{S_Q}}_i  \right) \left( \mathbf{Q}_i \mathbf{K}_j^T \right) \text{diag}\left( \mathbf{{S_K}}_j \right) \in \mathbb{R}^{B_r \times B_c} $.
        \State On chip, compute $m_i^{(j)} = \text{max}\left(m_i^{(j-1)}, \text{rowmax}\left(\mathbf{S}_i^{(j)}\right)\right) \in \mathbb{R}^{B_r}$. 
        \State On chip, compute $\mathbf{P}_i^{(j)} = \text{round}\left( R\times \text{exp}\left(\mathbf{S}_i^{(j)}-m_i^{(j)}\right) \right) \in  \mathbb{I}_8^{B_r \times B_c}$ (pointwise). 
        \State On chip, compute $l_i^{(j)} = l_i^{(j-1)}e^{m_i^{(j-1)}-m_i^{(j)}} + \text{rowsum}\left(\mathbf{P}_i^{(j)}\right) \in \mathbb{R}^{B_r}$.
        \State On chip, compute $\mathbf{O}_i^{(j)}=\text{diag}\left(e^{m_i^{(j-1)}-m_i^{(j)}}\right)
        \mathbf{O}_i^{(j-1)}+\mathbf{P}_i^{(j)}\mathbf{V}_j$
    \EndFor
    \State Load $\mathbf{S_V} \in \mathbb{R}$ from HBM to on-chip SRAM
    \State On chip, compute $\mathbf{O}_i=\text{diag}\left(l_i^{(T_c)}\right)^{-1}\mathbf{O}_i^{(T_c)}\mathbf{S_V}$.
    \State Write $\mathbf{O}_i$ to HBM as the $i$-th block of $\mathbf{O}$.
\EndFor
\State Return the output $\mathbf{O}$.
\end{algorithmic}
\end{algorithm}


\subsection{\aname{} Attention Design}
\label{subsec:superflash}

\aname{} implements the $\mathbf{Q}$, $\mathbf{K}$, and $\mathbf{V}$ matrices in self-attention module with fully INT8 format. 
In Figure~\ref{fig:alg_int8_attention}, we use green color to represent the data in INT8 format and the INT8 GEMM operations and use orange color to represent data and operations with FP32 format. 
The GEMM used in \aname{} takes two INT8 input matrices and produces an INT32 output matrix. 
As shown in Figure~\ref{fig:alg_int8_attention} and Algorithm~\ref{code:algo}, during inference workflow, all data stored in HBM ($\mathbf{Q}$, $\mathbf{K}$, and $\mathbf{V}$) and all matrix multiplication operations ($\mathbf{Q}_i \mathbf{K}_j^T$ and $\mathbf{P}_i^{(j)} \mathbf{V}_j$) are implemented with INT8 format.

We maintain two vector scalars $\mathbf{S_Q}$, $\mathbf{S_K} \in \mathbb{R}^{N}$, for token-level quantization of the $\mathbf{Q}$ and $\mathbf{K}$ matrices. For the $\mathbf{V}$ matrix, we maintain a constant scalar $\mathbf{S_V} \in \mathbb{R}$ for tensor-level quantization\footnote{We will implement $\mathbf{V}$ on a per-block quantization basis in our future work.}. 
After training, we perform linear symmetric quantization on the $\mathbf{Q}$, $\mathbf{K}$, and $\mathbf{V}$ matrices, quantizing them into INT8 format and obtaining scalars $\mathbf{S_Q} = \frac{rowmax(|\mathbf{Q}|)}{R}$, $\mathbf{S_K} = \frac{rowmax(|\mathbf{K}|)}{R}$, and $\mathbf{S_V} = \frac{max(|\mathbf{V}|)}{R}$, where $R=256$ is the INT8 quantization range.

As described in Algorithm~\ref{code:algo}, the attention forward workflow of \aname{} is similar to FlashAttention-2. 
Besides dividing $\mathbf{Q}$, $\mathbf{K}$, $\mathbf{V}$, and $\mathbf{O}$ into blocks (line 1-2), we also divide scalers $\mathbf{S_Q}$ and $\mathbf{S_K}$ into blocks (line 3). 
The outer loop iterates over each block $\mathbf{Q}_i$ of matrix $\mathbf{Q}$ (line 4-5).
For each block $\mathbf{Q}_i$, we also maintain the current row-wise maximum values $m_i^{(j)}$ and the sum of modified exponentials $l_i^{(j)}$ (explained later).
Both $m_i^{(j)}$ and $l_i^{(j)}$ are of FP32 format.

Figure~\ref{fig:alg_int8_attention} illustrates an iteration in the inner loop, which iterates over each block $\mathbf{K}_j$ and $\mathbf{V}_j$ of matrix $\mathbf{K}$ and $\mathbf{V}$ (line 7-8). 
We first call the integer GEMM kernel (with INT8 input and INT32 output) to multiply $\mathbf{Q}_i$ and $\mathbf{K}_j$, and then scale the result with $\mathbf{S_Q}_i$ and $\mathbf{S_K}_j$ to get the attention score matrix $\mathbf{S}_i^{(j)}$ with FP32 format (line 9). 
Thanks to the linearity of integer multiplication, the resulting attention score matrix $\mathbf{S}_i^{(j)}$ is equivalent to the standard attention score matrix obtained by first scaling $\mathbf{Q}_i$ and $\mathbf{K}_j$ with $\mathbf{S_Q}_i$ and $\mathbf{S_K}_j$, and then performing matrix multiplication with FP32 GEMM kernel on $ \left( \text{diag}\left( \mathbf{{S_Q}}_i  \right)  \mathbf{Q}_i \right) $ and $\left( \mathbf{K}_j^T  \text{diag}\left( \mathbf{{S_K}}_j \right) \right)$. 
We update row-wise maximum values $m^{(j)}$ with $\mathbf{S}_i^{(j)}$ (line 10). 

For the attention weight matrix $\mathbf{P}_i^{(j)}$, we aim to quantize it into INT8 format, allowing us to directly use INT8 GEMM for the $\mathbf{P}_i^{(j)}\mathbf{V}_j$ computation to update output (line 13).
According to the formulas in Section~\ref{subsec:softmax}, standard FlashAttention would compute $\tilde{\mathbf{P}}_i^{(j)} = \exp(\mathbf{S}_i^{(j)} - m_i^{(j)}) \in (0,1]^{B_r \times B_c}$ (line 11) and $l_i^{(j)} = l_i^{(j-1)}e^{m_i^{(j-1)}-m_i^{(j)}} + \text{rowsum}\left(\tilde{\mathbf{P}}_i^{(j)}\right) \in \mathbb{R}^{B_r}$ (line 12). 
As the values in $\tilde{\mathbf{P}}_i^{(j)}$ already falls between $(0, 1]$, we directly use $\mathbf{S_P} = \frac{1}{R}$ as the scaling factor for $\tilde{\mathbf{P}}_i^{(j)}$ and compute $\mathbf{P}_i^{(j)} = \text{round} \left( \tilde{\mathbf{P}}_i^{(j)}/\mathbf{S_P} \right)= \text{round}\left( R\times \text{exp}\left(\mathbf{S}_i^{(j)}-m_i^{(j)}\right) \right) \in  \mathbb{I}_8^{B_r \times B_c}$ (line 11).
Then we directly use the quantized weight matrix $\mathbf{P}_i^{(j)}$ to update the sum of expoentials $l_i^{(j)}$, thereby implicitly dividing the scaling factor $\mathbf{S_P}$ into $l_i^{(j)}$ (line 12). 
At the end of every iteration of the outer loop, we scale the final $\mathbf{O}_i^{(T_c)}$ with $\text{diag}\left(l_i^{(T_c)}\right)^{-1}$ to get the right output (line 16). 
In this process, we implicitly multiplied the scaling factor $\mathbf{S_P}$ into the output, completing the dequantization from integer to float.
Finally, we return the right output $\mathbf{O}$, which is equivalent to the output matrix calculated with standard softmax function and float computations.

Specifically, as in Section~\ref{subsec:softmax}, consider there are only one-row block and two column blocks in the attention matrix $\mathbf{S} = \left[\mathbf{S}^{(1)} \; \mathbf{S}^{(2)} \right]$ where $\mathbf{S}^{(1)}$, $\mathbf{S}^{(2)} \in \mathbb{R}^{B_r \times B_c}$. 
We have
\begin{align*}
m^{(1)} &= \text{rowmax}(\mathbf{S}^{(1)}) \in \mathbb{R}^{B_r} \\
\mathbf{P}^{(1)} &= \text{round}\left( R\times \text{exp}\left(\mathbf{S}^{(1)}-m^{(1)}\right) \right) \\ 
l^{(1)} &= \text{rowsum}(\mathbf{P}^{(1)})  \in \mathbb{R}^{B_r} \\
\widetilde{\mathbf{O}}^{(1)} &=  \mathbf{P}^{(1)} \mathbf{V}^{(1)} \in \mathbb{R}^{B_r \times d} \\
m^{(2)} &= \max(m^{(1)}, \text{rowmax}(\mathbf{S}^{(2)})) = m &\text{(Line 10)}\\
\mathbf{P}^{(2)} &= \text{round}\left( R\times \text{exp}\left(\mathbf{S}^{(2)}-m^{(2)}\right) \right)  &\text{(Line 11)}\\
l^{(2)} &= e^{m^{(1)} - m^{(2)}} l^{(1)} +  \text{rowsum}\left(\mathbf{P}^{(2)}\right) = R \times l &\text{(Line 12)} \\
\widetilde{\mathbf{O}}^{(2)} &= \text{diag}(e^{m^{(1)} - m^{(2)}}) \widetilde{\mathbf{O}}^{(1)} + \mathbf{P}^{(1)} \mathbf{V}^{(2)} = R \times e^{\mathbf{S}^{(1)} - m} \mathbf{V}^{(1)} + R \times e^{\mathbf{S}^{(2)} - m} \mathbf{V}^{(2)} &\text{(Line 13)}\\
\mathbf{O}^{(2)} &= \text{diag}(l^{(2)})^{-1} \widetilde{\mathbf{O}}^{(2)} = \frac{1}{R} \times \text{diag}(l)^{-1} \times (R \times e^{\mathbf{S}^{(1)} - m} \mathbf{V}^{(1)} + R \times e^{\mathbf{S}^{(2)} - m} \mathbf{V}^{(2)}) = \mathbf{O} &\text{(Line 16)}.
\end{align*}

Notice that in the above formulas, we neglect the small errors introduced by integer rounding.

\section{Empirical Evaluation}

We evaluate the inference speed and quantization accuracy of \aname{}, and compare it to FlashAttention with FP16 \cite{flashattention2} and FP8 \cite{flashattention3} data format. 
We also compare \aname{} with a half-INT8 version of \aname{} with INT8-format $\mathbf{Q}$ and $\mathbf{V}$ matrices and FP16-format $\mathbf{V}$ matrix. 
The experiments are conducted on an NVIDIA RTX4090 GPU. 
We highlight the following experimental results.

\bbb{Inference speed.}
\aname{} achieves about 72\% faster inference speed compared to FlashAttention with FP16 data format. 

\bbb{Quantization accuracy.}
\aname{} achieves about 46\% and 82\% smaller quantization error than FlashAttention with FP8 data format under normal-distributed and uniform-distributed activations, respectively.



\begin{figure}[tb]
    \centering
    \includegraphics[trim=-0.1cm 0cm -0.1cm 0cm, clip, width=1\linewidth]{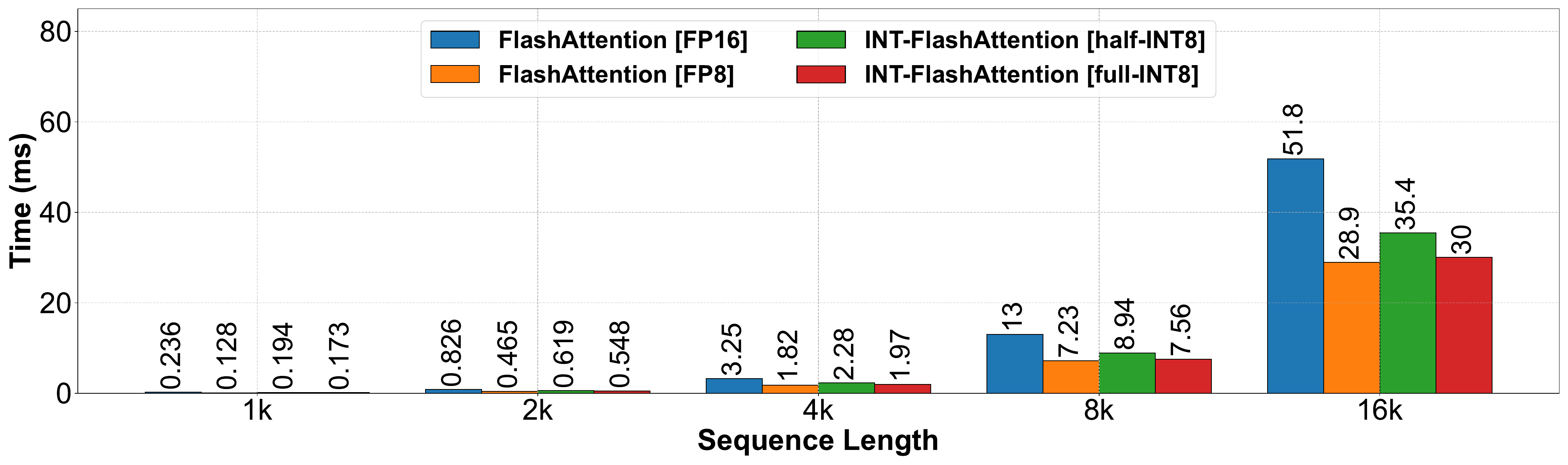}  
    \caption{Comparison of inference speed.}
    \label{fig:speed}
\end{figure}


\subsection{Inference Speed}
We implement \aname{} and the other candidate solutions in Triton, where we fix their batch size, number of heads, and dimension size per head to be the same. 
We evaluate the inference time with different context length. 
As shown in Figure~\ref{fig:speed}, compared to standard FlashAttention with FP16 data format, \aname{} achieve $31\%$, $52\%$, $66\%$, $72\%$, and $73\%$ smaller inference time under the sequence length of 1k, 2k, 4k, 8k, and 16k, respectively. 
We also notice that the inference time gap between \aname{} and FlashAttention-FP16 becomes larger as sequence length increases. 
On the other hand, \aname{} has nearly the same inference speed than FlashAttention with FP8 data format, with the gap narrowing as sequence length increases.



\subsection{Quantization Accuracy}

We compare the quantization accuracy of \aname{} with the other candidate solutions.
We manually create a one-layer self-attention module with the activations in its $\mathbf{Q}$, $\mathbf{K}$, and $\mathbf{V}$ matrices following normal distribution $\mathcal{N}(0,1)$ or uniform distribution $\mathcal{U}(-0.5, 0.5)$. 
We evaluate model quantization accuracy by measuring the Mean Relative Error (MRE) between original activations and activations after quantization and subsequent restoration. 
As shown in Table~\ref{tab:normal}, under normal-distributed activations, \aname{} achieves up to $1.8\times$ smaller MRE  compared to FlashAttention with FP8 data format.
As shown in Table~\ref{tab:uniform}, under uniform-distributed activations, \aname{} achieves up to $5.6\times$ smaller MRE compared to FlashAttention with FP8 data format.




\begin{table*}[h!]
\centering
\caption{Mean Relative Error (MRE) under normal-distributed activations.}
\label{tab:normal}
\begin{tabular}{c|c|c|c}
\toprule[1pt]
\textbf{\begin{tabular}[c]{@{}c@{}}Sequence \\ Length\end{tabular}} & \textbf{\begin{tabular}[c]{@{}c@{}}FlashAttention \\ {[}FP8{]}\end{tabular}} & \textbf{\begin{tabular}[c]{@{}c@{}} \aname{} \\ {[}half-INT8{]}\end{tabular}} & \textbf{\begin{tabular}[c]{@{}c@{}} \aname{} \\ {[}full-INT8{]}\end{tabular}} \\ \midrule[1pt]
1k  & 7.46\%   & 0.890\%  & 4.05\% \\
2k  & 7.50\%   & 0.802\%  & 4.18\%  \\
4k  & 7.66\%   & 0.843\%  & 4.21\%   \\
8k  & 7.51\%   & 0.932\%  & 4.38\%   \\
16k & 7.57\%   & 0.775\%  & 4.52\%   \\ 
\bottomrule[1pt]
\end{tabular}
\end{table*}


\begin{table*}[h!]
\centering
\caption{Mean Relative Error (MRE) under uniform-distributed activations.}
    \label{tab:uniform}
\begin{tabular}{c|c|c|c}
\toprule[1pt]
\textbf{\begin{tabular}[c]{@{}c@{}}Sequence \\ Length\end{tabular}} & \textbf{\begin{tabular}[c]{@{}c@{}}FlashAttention \\ {[}FP8{]}\end{tabular}} & \textbf{\begin{tabular}[c]{@{}c@{}}\aname{} \\ {[}half-INT8{]}\end{tabular}} & \textbf{\begin{tabular}[c]{@{}c@{}}\aname{} \\ {[}full-INT8{]}\end{tabular}} \\ \midrule[1pt]
1k  & 8.94\%  & 0.317\%  & 1.69\%  \\
2k  & 9.15\%  & 0.300\%  & 1.62\%  \\
4k  & 8.89\%  & 0.280\%  & 1.65\%  \\
8k  & 9.02\%  & 0.299\%  & 1.85\%  \\
16k & 8.97\%  & 0.296\%  & 1.82\%  \\ 
\bottomrule[1pt]
\end{tabular}
\end{table*}
\section{Conclusion, Limitations, and Future Work}

This paper proposes \aname{}, the first token-level INT8 post-training quantization architecture compatible with FlashAttention forward workflow. 
We implement a \aname{} prototype with fully INT8 activations and GEMMs, which significantly improves the inference speed of FlashAttention on Ampere GPUs. 
The limitation of this work is that the $\mathbf{V}$ matrix currently is only implemented with tensor-level quantization. 
It remains a significant challenge to quantize $\mathbf{V}$ on a per-token basis. In future work, we will use per-block quantization to optimize the implementation of the $\mathbf{V}$ matrix. 
We also plan to combine our \aname{} with Hadamard transformations to further accelerate the inference process while maintaining high accuracy of the weight coefficients.

\bibliography{iclr2024_conference}

\begin{thebibliography}{26}
\providecommand{\natexlab}[1]{#1}
\providecommand{\url}[1]{\texttt{#1}}
\expandafter\ifx\csname urlstyle\endcsname\relax
  \providecommand{\doi}[1]{doi: #1}\else
  \providecommand{\doi}{doi: \begingroup \urlstyle{rm}\Url}\fi

\bibitem[Achiam et~al.(2023)Achiam, Adler, Agarwal, Ahmad, Akkaya, Aleman, Almeida, Altenschmidt, Altman, Anadkat, et~al.]{achiam2023gpt}
Josh Achiam, Steven Adler, Sandhini Agarwal, Lama Ahmad, Ilge Akkaya, Florencia~Leoni Aleman, Diogo Almeida, Janko Altenschmidt, Sam Altman, Shyamal Anadkat, et~al.
\newblock Gpt-4 technical report.
\newblock \emph{arXiv preprint arXiv:2303.08774}, 2023.

\bibitem[Chen et~al.(2024)Chen, Li, Xu, Zhu, Li, Tian, Barsoum, Wang, and Cheng]{chen2024ternaryllm}
Tianqi Chen, Zhe Li, Weixiang Xu, Zeyu Zhu, Dong Li, Lu~Tian, Emad Barsoum, Peisong Wang, and Jian Cheng.
\newblock Ternaryllm: Ternarized large language model.
\newblock \emph{arXiv preprint arXiv:2406.07177}, 2024.

\bibitem[Dally(2015)]{dally2015high}
William Dally.
\newblock High-performance hardware for machine learning.
\newblock \emph{Nips Tutorial}, 2:\penalty0 3, 2015.

\bibitem[Dao(2023)]{flashattention2}
Tri Dao.
\newblock Flashattention-2: Faster attention with better parallelism and work partitioning.
\newblock \emph{arXiv preprint arXiv:2307.08691}, 2023.

\bibitem[Dao et~al.(2022)Dao, Fu, Ermon, Rudra, and R{\'e}]{flashattention1}
Tri Dao, Dan Fu, Stefano Ermon, Atri Rudra, and Christopher R{\'e}.
\newblock Flashattention: Fast and memory-efficient exact attention with io-awareness.
\newblock \emph{Advances in Neural Information Processing Systems}, 35:\penalty0 16344--16359, 2022.

\bibitem[Dettmers et~al.(2022)Dettmers, Lewis, Belkada, and Zettlemoyer]{dettmers2022gpt3}
Tim Dettmers, Mike Lewis, Younes Belkada, and Luke Zettlemoyer.
\newblock Gpt3. int8 (): 8-bit matrix multiplication for transformers at scale.
\newblock \emph{Advances in Neural Information Processing Systems}, 35:\penalty0 30318--30332, 2022.

\bibitem[Dettmers et~al.(2023)Dettmers, Svirschevski, Egiazarian, Kuznedelev, Frantar, Ashkboos, Borzunov, Hoefler, and Alistarh]{dettmers2023spqr}
Tim Dettmers, Ruslan Svirschevski, Vage Egiazarian, Denis Kuznedelev, Elias Frantar, Saleh Ashkboos, Alexander Borzunov, Torsten Hoefler, and Dan Alistarh.
\newblock Spqr: A sparse-quantized representation for near-lossless llm weight compression.
\newblock \emph{arXiv preprint arXiv:2306.03078}, 2023.

\bibitem[Frantar \& Alistarh(2022)Frantar and Alistarh]{frantar2022optimal}
Elias Frantar and Dan Alistarh.
\newblock Optimal brain compression: A framework for accurate post-training quantization and pruning.
\newblock \emph{Advances in Neural Information Processing Systems}, 35:\penalty0 4475--4488, 2022.

\bibitem[Frantar et~al.(2022)Frantar, Ashkboos, Hoefler, and Alistarh]{frantar2022gptq}
Elias Frantar, Saleh Ashkboos, Torsten Hoefler, and Dan Alistarh.
\newblock Gptq: Accurate post-training quantization for generative pre-trained transformers.
\newblock \emph{arXiv preprint arXiv:2210.17323}, 2022.

\bibitem[Huang et~al.(2024)Huang, Liu, Qin, Li, Zhang, Liu, Magno, and Qi]{huang2024billm}
Wei Huang, Yangdong Liu, Haotong Qin, Ying Li, Shiming Zhang, Xianglong Liu, Michele Magno, and Xiaojuan Qi.
\newblock Billm: Pushing the limit of post-training quantization for llms.
\newblock \emph{arXiv preprint arXiv:2402.04291}, 2024.

\bibitem[Lee et~al.(2023)Lee, Jin, Kim, Kim, and Park]{lee2023owq}
Changhun Lee, Jungyu Jin, Taesu Kim, Hyungjun Kim, and Eunhyeok Park.
\newblock Owq: Lessons learned from activation outliers for weight quantization in large language models.
\newblock \emph{arXiv preprint arXiv:2306.02272}, 2, 2023.

\bibitem[Lee et~al.(2024)Lee, Bae, Kim, Kwon, and Lee]{lee2024fp8}
Joonhyung Lee, Jeongin Bae, Byeongwook Kim, Se~Jung Kwon, and Dongsoo Lee.
\newblock To fp8 and back again: Quantifying the effects of reducing precision on llm training stability.
\newblock \emph{arXiv preprint arXiv:2405.18710}, 2024.

\bibitem[Li et~al.(2024)Li, Ning, Wang, Liu, Shi, Yan, Dai, Yang, and Wang]{li2024evaluating}
Shiyao Li, Xuefei Ning, Luning Wang, Tengxuan Liu, Xiangsheng Shi, Shengen Yan, Guohao Dai, Huazhong Yang, and Yu~Wang.
\newblock Evaluating quantized large language models.
\newblock \emph{arXiv preprint arXiv:2402.18158}, 2024.

\bibitem[Li et~al.(2021)Li, Gong, Tan, Yang, Hu, Zhang, Yu, Wang, and Gu]{li2021brecq}
Yuhang Li, Ruihao Gong, Xu~Tan, Yang Yang, Peng Hu, Qi~Zhang, Fengwei Yu, Wei Wang, and Shi Gu.
\newblock Brecq: Pushing the limit of post-training quantization by block reconstruction.
\newblock \emph{arXiv preprint arXiv:2102.05426}, 2021.

\bibitem[Micikevicius et~al.(2022)Micikevicius, Stosic, Burgess, Cornea, Dubey, Grisenthwaite, Ha, Heinecke, Judd, Kamalu, et~al.]{micikevicius2022fp8}
Paulius Micikevicius, Dusan Stosic, Neil Burgess, Marius Cornea, Pradeep Dubey, Richard Grisenthwaite, Sangwon Ha, Alexander Heinecke, Patrick Judd, John Kamalu, et~al.
\newblock Fp8 formats for deep learning.
\newblock \emph{arXiv preprint arXiv:2209.05433}, 2022.

\bibitem[Milakov \& Gimelshein(2018)Milakov and Gimelshein]{milakov2018online}
Maxim Milakov and Natalia Gimelshein.
\newblock Online normalizer calculation for softmax.
\newblock \emph{arXiv preprint arXiv:1805.02867}, 2018.

\bibitem[Morgan(2024)]{ampere}
Timothy~Prickett Morgan.
\newblock Top500 supers: This is peak nvidia for accelerated supercomputers.
\newblock 2024.
\newblock URL \url{https://www.nextplatform.com/2024/05/13/top500-supers-this-is-peak-nvidia-for-accelerated-supercomputers/}.

\bibitem[Rabe \& Staats(2021)Rabe and Staats]{rabe2021self}
Markus~N Rabe and Charles Staats.
\newblock Self-attention does not need $ o (n^{2}) $ memory.
\newblock \emph{arXiv preprint arXiv:2112.05682}, 2021.

\bibitem[Shah et~al.(2024)Shah, Bikshandi, Zhang, Thakkar, Ramani, and Dao]{flashattention3}
Jay Shah, Ganesh Bikshandi, Ying Zhang, Vijay Thakkar, Pradeep Ramani, and Tri Dao.
\newblock Flashattention-3: Fast and accurate attention with asynchrony and low-precision.
\newblock \emph{arXiv preprint arXiv:2407.08608}, 2024.

\bibitem[Shang et~al.(2023)Shang, Yuan, Wu, and Dong]{shang2023pb}
Yuzhang Shang, Zhihang Yuan, Qiang Wu, and Zhen Dong.
\newblock Pb-llm: Partially binarized large language models.
\newblock \emph{arXiv preprint arXiv:2310.00034}, 2023.

\bibitem[Tao et~al.(2022)Tao, Hou, Zhang, Shang, Jiang, Liu, Luo, and Wong]{tao2022compression}
Chaofan Tao, Lu~Hou, Wei Zhang, Lifeng Shang, Xin Jiang, Qun Liu, Ping Luo, and Ngai Wong.
\newblock Compression of generative pre-trained language models via quantization.
\newblock \emph{arXiv preprint arXiv:2203.10705}, 2022.

\bibitem[Touvron et~al.(2023)Touvron, Martin, Stone, Albert, Almahairi, Babaei, Bashlykov, Batra, Bhargava, Bhosale, et~al.]{touvron2023llama}
Hugo Touvron, Louis Martin, Kevin Stone, Peter Albert, Amjad Almahairi, Yasmine Babaei, Nikolay Bashlykov, Soumya Batra, Prajjwal Bhargava, Shruti Bhosale, et~al.
\newblock Llama 2: Open foundation and fine-tuned chat models.
\newblock \emph{arXiv preprint arXiv:2307.09288}, 2023.

\bibitem[van Baalen et~al.(2023)van Baalen, Kuzmin, Nair, Ren, Mahurin, Patel, Subramanian, Lee, Nagel, Soriaga, et~al.]{van2023fp8}
Mart van Baalen, Andrey Kuzmin, Suparna~S Nair, Yuwei Ren, Eric Mahurin, Chirag Patel, Sundar Subramanian, Sanghyuk Lee, Markus Nagel, Joseph Soriaga, et~al.
\newblock Fp8 versus int8 for efficient deep learning inference.
\newblock \emph{arXiv preprint arXiv:2303.17951}, 2023.

\bibitem[Vaswani et~al.(2017)Vaswani, Shazeer, Parmar, Uszkoreit, Jones, Gomez, Kaiser, and Polosukhin]{vaswani2017attention}
Ashish Vaswani, Noam Shazeer, Niki Parmar, Jakob Uszkoreit, Llion Jones, Aidan~N Gomez, Łukasz Kaiser, and Illia Polosukhin.
\newblock Attention is all you need.
\newblock \emph{Advances in Neural Information Processing Systems}, 2017.

\bibitem[Xu et~al.(2024)Xu, Zhang, Yang, Liu, Huang, Xu, and Liu]{xu2024empowering}
Daliang Xu, Hao Zhang, Liming Yang, Ruiqi Liu, Gang Huang, Mengwei Xu, and Xuanzhe Liu.
\newblock Empowering 1000 tokens/second on-device llm prefilling with mllm-npu.
\newblock \emph{arXiv preprint arXiv:2407.05858}, 2024.

\bibitem[Zhou et~al.(2024)Zhou, Ning, Hong, Fu, Xu, Li, Lou, Wang, Yuan, Li, et~al.]{zhou2024survey}
Zixuan Zhou, Xuefei Ning, Ke~Hong, Tianyu Fu, Jiaming Xu, Shiyao Li, Yuming Lou, Luning Wang, Zhihang Yuan, Xiuhong Li, et~al.
\newblock A survey on efficient inference for large language models.
\newblock \emph{arXiv preprint arXiv:2404.14294}, 2024.

\end{thebibliography}
\bibliographystyle{iclr2024_conference}

\end{document}